# Impact of translation on biomedical information extraction from real-life clinical notes

Christel Gérardin[a,*], Yuhan Xiong[a,b], Perceval Wajsbürt[c], Fabrice Carrat[a,d], Xavier Tannier[e]

[a]IPLESP, 27 rue de Chaligny, Paris, 75012, France

[b]Shanghai Jiaotong University, 800 Dongchuan RD. Minhang District, Shanghai, China

[c]Innovation and Data unit, IT Department, Assistance Publique- hôpitaux de Paris, 33 bd Picpus, Paris, 75012, France

[d]Public Health Department, Hôpital Saint-Antoine, Assistance Publique- hôpitaux de Paris, 184 Rue du Faubourg Saint-Antoine, Paris, 75012, France

[e]Sorbonne Université, Inserm, Université Sorbonne Paris Nord, Laboratoire d'Informatique Médicale et d'Ingénierie et des Connaissances en e-Santé, LIMICS, 15 rue de l'école de médecine, Paris, F-75006, France

*Corresponding author : christel.ducroz-gerardin@iplesp.upmc.fr

## Abstract

The objective of our study is to determine whether using English tools to extract and normalize French medical concepts on translations provides comparable performance to French models trained on a set of annotated French clinical notes.

We compare two methods: a method involving French language models and a method involving English language models. For the native French method, the Named Entity Recognition (NER) and normalization steps are performed separately. For the translated English method, after the first translation step, we compare a two-step method and a terminology-oriented method that performs extraction and normalization at the same time. We used French, English and bilingual annotated datasets to evaluate all steps (NER, normalization and translation) of our algorithms.

Concerning the results, the native French method performs better than the translated English one with a global f1 score of 0.51 [0.47;0.55] against 0.39 [0.34;0.44] and 0.38 [0.36;0.40] for the two English methods tested.

In conclusion, despite the recent improvement of the translation models, there is a significant performance difference between the two approaches in favor of the native French method which is more efficient on French medical texts, even with few annotated documents.

## 1. Introduction

Named Entity Recognition (NER) and term normalization are important steps in biomedical Natural Language Processing (NLP). NER is used to extract key information from textual medical reports and normalization consists of mapping a specific term to its formal reference in a shared terminology such as UMLS® [1]. Major improvements have been made recently in these areas, especially in English, as a huge amount of data is available in the literature and resources. Modern automatic language processing relies heavily on pre-trained language models, which allow for efficient semantic representation of texts. The development of algorithms such as transformers [2, 3] has led to significant progress in this area.

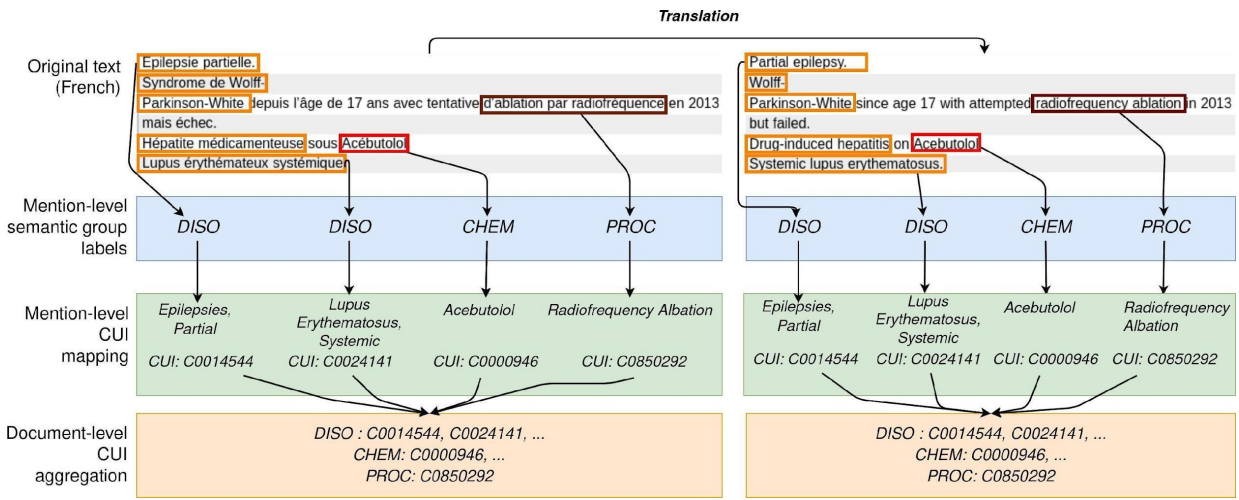

**Figure 1: Overall method objective: from raw text to CUI information document by document.** The "Mention level" denotes the analysis done at the word or group of words level: first, the NER step (in blue), then the normalization (in green), finally all mentions with normalized CUIs are aggregated at the document level (orange part). Both native French and translated English approaches are compared.

In many languages other than English, efforts still need to be made to obtain such interesting results, in particular due to a much smaller amount of accessible data [4].

In this context, our work explores the question of the relevance of a translation step for the recognition and normalization of medical concepts in biomedical documents in French. We compare two methods: 1) a native French approach were only annotated documents and resources in French are used, and 2) a translation-based approach where documents are translated into English, in order to take advantage of existing tools and resources for this language which would allow to extract concepts mentioned in unseen French texts without new training data (zero-shot) as proposed in Van Mulligen et al. [5].

We evaluate and discuss the results on several French biomedical corpora, including a new set of 42 hospitalization reports annotated with 4 entity groups. We evaluate the normalization task at the document-level, in order to avoid a cross-lingual alignment step at evaluation time, which would add a potential level of error and thus would make the results more difficult to interpret (see word alignment in [6, 7]). This normalization is performed by matching all the terms to their *Concept Unique Identifier* (CUI) in the UMLS® [1]. Figure 1 summarizes these different steps, from the raw French text and the translated English text to CUI aggregation and comparison at document-level. All our codes are available on github [8].

## 2. Background

The different steps of our algorithms rely heavily on Transformers language models [2]. These models are currently the state of the art for many natural language processing (NLP) tasks, such as machine translation, named entity recognition, classification, and text normalization (also known as "entity binding"). Once trained, these models can represent any specific language, such as biomedical language or legal language. The power of these models comes from their neural architecture but also depends largely on the amount of data on which they are trained. In the biomedical domain, two main types of data are available: public articles (e.g. PubMed ) and clinical electronic medical records databases (e.g. MIMIC III [14]), and the most powerful models are for example BioBERT [15] which has been trained on the whole of PubMed in English, and ClinicalBERT [16] trained on PubMed and MIMIC III. In French language, the variety of models is less important, with the models CamemBERT [17] and FlauBERT [18] for the general domain, and no specific model available for the biomedical domain.

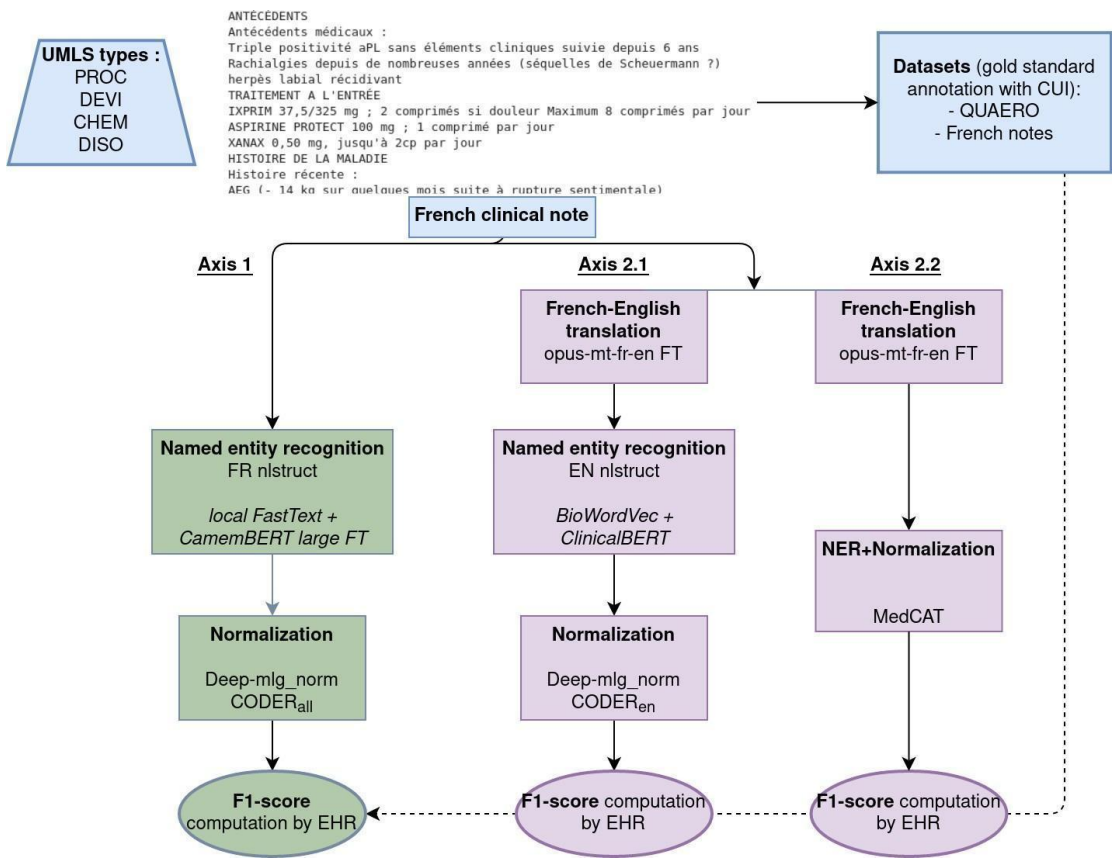

**Figure 2: Diagram of the different experiments comparing French and English language models without and with intermediate translation steps.** The Axis 1 (green axis on the left) corresponds to the native French branch with a NER step based on a FastText model trained from scratch on French clinical notes and a CamemBERT model. A multilingual BERT model is then used for the normalization step with two models tested : a deep multilingual normalization model [9] and CODER[10] with the *all* version. The Axes 2.1 and 2.2 (two purple axes on the right) correspond to translated English branches with a first translation step done by opus-mt-fr-en model[11] for both. Axis 2.1 (on the left) with decoupled NER and normalization steps, based on FastText trained from PubMed and Mimic III [12] for the NER part, and deep multilingual normalization[9] or CODER[10] with the *English* version, for the normalization. Axis 2.2 (on the right) uses a single system for both NER and normalization steps: MedCAT [13].

In addition to the particularly powerful English pre-trained models, universal biomedical terminologies (i.e., metathesaurus) also contain significantly more English terms than other languages. For example, the Unified Medical Language System (UMLS®[1]) contains at least ten times more English terms than French terms, which can enable ruled-based models to perform better in English. As mentioned above, each concept of reference in the UMLS®[1] is assigned a Concept Unique Identifier (CUI), associated with a set of synonyms eventually in several languages, and a semantic group -such as *Disorders, Chemical and Drugs, Procedure, Anatomy* and so on.

At the same time, machine translation has also gained in performance thanks to the same type of language models based on transformers, and the last few years have seen the emergence of high-quality automatic translation such as opus-mt developed by Tiedemann et al. [11], Google Translate® and others. These last two observations led several research teams to add a translation step in order to analyze medical texts, for instance to extract relevant mentions in ultrasound reports [19, 20] or in the case of medical concept normalization [9, 10, 21]. Work in the general (non-medical) domain has also focused on alignment between named entities in parallel bilingual texts [22, 23].

## 3. Materials and Methods

### 3.1. Overview

Figure 2 presents the main approaches and models used in our study. We explored a "native French approach axis" (axis 1 in Figure 2) based on French language models learned on and applied to French annotated data; and two "translated English approach axes" (axes 2.1 and 2.2) based on a translation step and English concept extraction tools. We compare the performance of all axes with an average of the CUI predictions accuracies at document level, for all documents.

### 3.1.1. Native French approach

Axis 1 consists of two steps: a NER step and a normalization step. For the NER step, we used the nested named entity extraction algorithm presented in the Section 3.4 below. Then, a normalization step is performed by two different algorithms : a deep multilingual normalization model [9] and CODER [10] with the *all* version (detailed in sections 3.5.1 and 3.5.2 respectively).

### 3.1.2. Translated-English approach

Axes 2.1 and 2.2 first consist in a translation step, presented in section 3.3 below, operated by the state-of-the-art *opus-mt-fr-en* algorithm [11] or Google Translate®. Then, like axis 1, axis 2.1 is based on a NER and a Normalization step. The NER step is done by the same algorithm but trained on the n2c2 2019 dataset[24], for the normalization step with used the same deep multilingual algorithm[9] (section 3.5.1) and the English version of CODER [10] based on a BioBERT[15] model. This axis allows to compare two methods whose difference is only the translation step.

Axis 2.2 is based on the MedCAT[13] algorithm presented in section 3.5.3 which performs the NER and normalization simultaneously. In this case, we compare the native French method with a state-of-the-art English system, ready to use, which is not available in French.

## 3.2. Datasets

### 3.2.1. Overview

For all our experiments, we chose to focus on only four UMLS®[1] group: *Chemical & Drugs* (CHEM) and *Devices* (DEVI) corresponding to medical devices such as pace-maker, catheter, etc., *Disorders* (DISO) corresponding to all signs, symptoms, findings (for instance positive or negative biological test results) and diseases, *Procedures* (PROC) correspond- ing to all diagnostic and therapeutic procedures such as imaging, biological tests, operative procedures, etc.

Table 1 presents the datasets used for all our experiments with the corresponding numbers of documents. First, two French datasets were used for the final evaluation, as well as for training the axis-1 models. QUAERO is a freely available corpus [25] based on pharmacological notes with two sub-corpora: Medline (PubMed abstract short sentences) and EMEA (drug notices). We also annotated a new real-life clinical notes dataset from the Assistance-Publique Hôpitaux de Paris datawarehouse, described in Supplementary materials Section 1.1.1.

Second, we used the corpus n2c2 2019 [24] with annotated CUIs – on which we auto- matically added the UMLS®[1] semantic group information,  to train the axis-2.1 system and to evaluate the English NER and normalization algorithms. We also used the Mantra dataset [26], corresponding to a multilingual gold standard corpus for biomedical concept recognition.

Finally, we fine-tuned and tested the translation algorithms on both 2016 [27] and 2019 [28] WMT biomedical corpora. Detailed description of the number of respective entities in the datasets can be found in the supplementary Table 1.

Table 1: **Datasets**. Presentation of all datasets used.

| Language | French | | | English | | English-French | |
|---|---|---|---|---|---|---|---|
| Datasets | Quaero | | French notes | n2c2_2019 | Mantra (English) | WMT 2016 | WMT 2019 |
| | EMEA | Medline | | | | | |
| Type | Drug notices | Medline titles | French notes | English notes | Drug not. & Medline titles | Pubmed abstracts | Pubmed abstracts |
| Size (docs) | 38 | 2514 | 42 | 100 | 200 | > 600k sent. | 6542 |
| Used for : | | | | | | | |
| Train NER | x | x | x | x | | | |
| Test NER | x | x | x | x | | | |
| Normalisation | x | x | x | x | | | |
| Test MedCAT | | | | x | x | | |
| Translation (Fine Tuning) | | | | | | x | x |
| Translation (test) | | | | | | x | |

French corpus annotation methods are detailed in section 1.1.1 of supplementary materials with supplementary Figure 1. Entities repartition for this annotation is detailed in Supplementary Table 1.

### 3.3. Translation

We used and compared two main algorithms for the translation step: the opus-mt-fr-en model [11] that we tested without and with *fine-tuning* on the two biomedical translation corpora from 2016 and 2019 [27, 28], and Google Translate® as a comparison model.

### 3.4. Named Entity Recognition

For this step, we used the algorithm of Wajsburt et al.[29] described in [30]. This model is based on the representation of a BERT transformer [3] and computes the scores of all possible concepts to be predicted in the text. The extracted concepts are delimited by three values: (start, end, label). More precisely, the text encoding corresponds to the last 4 layers of BERT, the Fasttext embedding and a Char-CNN max-pool representation [31] of the word. The decoding step is then performed by a 3-layer LSTM [32] with learnable gating weights [33], similar to the method in [34]. A sigmoid function is added to the top. Values (start, end, label) with a score greater than 0.5 are retained for prediction. The loss function is a binary cross-entropy and we used the Adam optimizer [35].

In our experiments, for the native French axis (axis 1 on Figure 2), the pre-trained embeddings used to train the model were based on a FastText [36] trained from scratch on 5 Gigabytes of clinical text and a camemBERT-large [17] *fine-tuned* on this same dataset. For the English axis 2.1, the pre-trained models were BioWordVec [12] and clinicalBERT [16].

### 3.5. Normalization algorithms

This step of our experiments is essential in order to compare a native French and a translated English method and consists in mapping each mention extracted from the text to its associated CUI in the UMLS®[1]. We compare three models for this step, described below: the deep multilingual normalization algorithm developed by [9], the CODER[10] and the MedCAT model[13] that performs both NER and normalization at the same time.

All these three models do not need any training dataset other than the UMLS®.

### 3.5.1. Deep multilingual normalization

This algorithm from Wajsbürt et al. [9] considers the normalization task as a highly-multiclass classification problem with a cosine similarity and a softmax function as a last layer.

The model is based on contextualized embedding, using the pre-trained multilingual BERT model [3] and works in two steps: during the first step, the BERT model is *fine-tuned* and French UMLS terms and their corresponding English synonyms are learned. Then, in the second step, the BERT model is frozen and the representation of all English-only terms (i.e. only present in English in the UMLS®[1]) is learned. The same training is used for the native French and translated English approach. This model was trained with the 2021 UMLS®[1] version, corresponding to the version used for the annotation of the French corpus. This model was thus trained on more than 4 million concepts corresponding to 2 million CUIs.

### 3.5.2. CODER

The CODER algorithm [10] is developed through contrastive learning based on the UMLS®[1] medical knowledge graph, concept similarities are computing from terms representation and relation of this knowledge graph. The contrastive learning is used to learn embeddings through a Multi-Similarity loss [37]. The authors have developed two versions: a multilingual based on multilingual BERT [3] and an English one based on BioBERT[15] pre-trained model. We used the multilingual version for axis 1 (native French approach), and the English version for axis 2.1. The two types of this model (CODER all and CODER en) were trained with the 2020 UMLS version (publicly available models). The CODER all [10] was thus trained on more than 4 million concepts corresponding to 2 million CUIs and the CODER en was trained on more than 3 millions terms and 2 millions CUI.

For the deep multilingual model and the CODER model, in order to improve performances in terms of accuracy, we chose to add the semantic group information (*i.e.* CHEM, DEVI, DISO, PROC) to the output of the model: namely, among the *k* first CUIs chosen from a mention, we choose the first one of the right group.

The MedCAT algorithm is detailed in Section 1.1.1 (supplementary materials).

Table 2: **NER performance.** Results of the NER models. For all experiments we used the same NER algorithm described in section 3.4, but with different pre-trained models. FastText* corresponds to a FastText [36] trained from scratch on our local clinical dataset.

| Data set | | EMEA test | | | French notes | | | n2c2 2019 test | | |
|---|---|---|---|---|---|---|---|---|---|---|
| Models | | FastText* & camemBERT-FT | | | FastText* & camemBERT-FT | | | BioWordVec [12] & ClinicalBERT [16] | | |
| | | preci-sion | recall | f1-score | preci-sion | recall | f1-score | preci-sion | recall | f1-score |
| Groups | CHEM | 0.80 | 0.83 | 0.82 | 0.84 | 0.88 | 0.86 | 0.87 | 0.85 | 0.86 |
| | DEVI | 0.42 | 0.81 | 0.55 | 0.00 | 0.00 | 0.00 | 0.58 | 0.51 | 0.54 |
| | DISO | 0.54 | 0.63 | 0.59 | 0.67 | 0.65 | 0.66 | 0.74 | 0.72 | 0.73 |
| | PROC | 0.73 | 0.78 | 0.74 | 0.78 | 0.72 | 0.75 | 0.80 | 0.78 | 0.79 |
| | **Overall** | **0.71** | **0.77** | **0.74** | **0.73** | **0.71** | **0.72** | **0.78** | **0.76** | **0.77** |

## 4. Results

The sections below present the performance results for each step. The n2c2 2019 challenge corpus [24] allowed us to evaluate the performance of our English models on clinical data and the Biomedical Translation shared task 2016 [27] to evaluate our translation performance on biomedical data with a BLEU score [38].

### 4.1. NER performances

To be able to compare our native French and translated English approaches, we used the same

NER model (section 3.4), trained and tested on each respective datasets described above (section 3.2). Table 2 presents the corresponding results. The overall F1-scores are similar from one dataset to another: from 0.72 to 0.77.

## 4.2. Normalization performance

This section only exposes the normalization performance based on the gold standard entity mentions, without the intermediate steps. The results are summarized in Table 3. The deep multilingual algorithm performs better for all tested corpora, with an improvement in F1 score from +0.6 to +0.11. For comparison, the winning team of the 2019 n2c2 challenge had achieved an accuracy of 0.85 using the n2c2 dataset directly to train their algorithm [24]. In our context of comparing algorithms between two languages, the normalization algorithms are not trained on data other than UMLS®. The performance of MedCAT (presented in Supplementary Table 2) cannot be directly compared to other models since this method performs both NER and normalization in one step. However, we find that this algorithm performs as well as Axis 2.1 for overall performance, as shown in Table 5.

Table 3: **Normalization performance.** Presentation of the accuracy results of the Normalization models computed from the annotated datasets, focusing on the four semantic groups of interest : CHEM, DEVI, DISO, PROC.

| Dataset<br>Models | EMEA test | French notes | n2c2 2019 test |
|---|---|---|---|
| deep mlg norm | **0.65** | **0.57** | **0.74** |
| CODER all | 0.58 | 0.51 | – |
| CODER en | – | – | 0.63 |

### 4.3. Translation performances

For the two translation models, the respective BLEU scores [38] are computed on the 2016 Biomedical Translation shared task [27]. A fine-tuned version of opus-mt-fr-en [11] on the 2016 and 2019 Biomedical Translation shared tasks was also tested. However, the Google translate model could not be used for our experiments involving clinical notes due to confidentiality reasons.

Table 4 shows the BLEU score results for the three models, showing that fine-tuning on the opus-mt-fr-en model [11] on biomedical datasets led to the best results, with a BLEU score[38] of 0.51. We will use this model for the overall performance of axes 2.1 and 2.2.

### 4.4. Overall performances from raw text to CUI predictions

This section presents the overall performance of the 3 axes, in an end-to-end pipeline. For axis 2, the results are those obtained with the best normalization algorithm (presented in Table 3). The model used for translation was the opus-mt-fr-en [11] fine-tuned model. The results are presented in Table 5, the best results are obtained by the native French approach on the EMEA corpus [25] and the French clinical notes. The 95% confidence intervals were calculated using the empirical bootstrap method [39].

Table 4: **Translation performances.** BLEU scores of Translation models. *opus-mt-fr-en* FT corresponds to the *opus-mt-fr-en* model [11] *fine-tuned* on biomedical translated corpus from [27] and [28].

| Data set | | wmt biomed 2016 test |
|---|---|---|
| Models | Google Translate | 0.42 |
| | opus-mt-fr-en | 0.31 |
| | opus-mt-fr-en FT | **0.51** |

Table 5: **Overall performances.** The normalization step is performed by the deep multilingual model and the translation by the opus-mt-fr-en FT model.

| | | EMEA test | | | French notes | | |
|---|---|---|---|---|---|---|---|
| | | precision | recall | f1-score | precision | recall | f1-score |
| Methods | Axis 1 (French NER+normalization) | 0.63 | 0.60 | **0.61** **[0.53;0.65]** | 0.49 | 0.53 | **0.51** **[0.47;0.55]** |
| | Axis 2.1 (Translation+NER+normalization) | 0.53 | 0.40 | 0.45 [0.38;0.51] | 0.41 | 0.38 | 0.39 [0.34;0.44] |
| | Axis 2.2 (Translation+MedCAT[13]) | 0.53 | 0.46 | 0.49 [0.38;0.54] | 0.38 | 0.38 | 0.38 [0.36;0.40] |

## 5. Discussion

In this paper, we compared two approaches for extracting medical concepts from clinical notes. A French approach based on a French language model and a translated English approach where we compare two state-of-the-art English biomedical language models, after a translation step. The main advantages of our experiment are that it is reproducible, and that we were able to analyze the performance of each step of the algorithm: NER, normalization and translation, and to test several models for each step.

### 5.1.1. The quality of the translation is not sufficient

We show that the native French approach outperforms the two translated English approaches, even with a small French training dataset. This analysis confirms that, when possible, an annotated dataset improves feature extraction. The evaluation of each intermediate step allows us to show that the performance of each module is similar in French and in English. We can then conclude that it is rather the translation phase itself that is of insufficient quality to allow the use of English as a proxy without loss of performance. This is confirmed by the performance calculations of the translation, where the calculated BLEU scores are relatively low, although improved by a fine-tuning step.

In conclusion, although translation is commonly used for entity extraction or term normalization in languages other than English [20, 40, 41, 42, 5], due to the availability of turnkey models that do not require additional annotation by a clinician, we show that this induces a significant performance loss.

Commercial API-based translation services could not be used for our task due to data privacy issues. However, the opus-mt model is considered state of the art, it is adjustable on domain specific data, and the translation results presented in Table 4 confirm the lack of performance difference between this model and the google translate model.

Even if our experiments were performed on only one language, the French-English pair is one of the best performing in recent translation benchmarks[43]. It is unlikely that other languages would lead to significantly better results.

*5.1.2. Error Analysis*

In these experiments, the overall results may appear low, but the task is still complex, especially because the UMLS® [1] contains many synonyms with different CUIs. To better understand, we performed an error analysis on the normalization task only, as shown in Supplementary Table 3, with a physician's evaluation, on a sample of 100 errors for both models. We calculated that 24% and 39% of the terms found by the deep normalization algorithm [9] and CODER [10] respectively were actually synonyms but with two different UMLS CUIs. For example, cardiac ultrasound has CUI C1655737 while echocardiography has another CUI

C0013516, similarly H/O: thromboembolism has a CUI of C0455533 while history of thromboembolism has a CUI of C1997787 and so on. In addition, as shown in Supplementary Table 3, abbreviations and misspelled words also induce many errors and are difficult to manage, even though some abbreviations are already built into UMLS. Another limitation comes from the ever-changing versions of the UMLS®. In any case, it is the relative differences between the results that matter for our purposes, not the absolute values.

### 5.1.3. Limitations

This work has several limitations, first of all, the real-life French clinical notes had very few terms attached to the "Devices" semantic group, thus preventing the NER algorithm from finding them in the test dataset. However, this drawback, penalizing the native French approach, still allows us to conclude on the results. Moreover, in this study, we did not take into account the attributes of the extracted terms such as the negation, the hypothetical attribute or the belonging to another person than the patient, this for comparison purposes, indeed the datasets QUAERO [25] and n2c2 2019 [24] did not have this information labeled.

**Ethics**

The study and its experimental protocol was approved by the AP-HP Scientific and Ethical Committee (IRB00011591 decision number CSE 20-0093). Patients were informed that their EHR information could be reused after an anonymization process and those who objected to the reuse of their data were excluded. All methods were carried out in accordance with relevant guidelines (reference methodology MR-004 of the CNIL: Commission Nationale de l'Informatique et des Libertés [44]).

**Data availability**

The datasets analyzed during the current study are not publicly available due the confidentiality of data from patient records, even after de-identification. However, access to the AP-HP data warehouse's raw data can be granted following the process described on its website: www.eds.aphp.fr, contacting the Ethical and Scientific Commity at secretariat.cse@aphp.fr. A prior validation of the access by the local institutional review board is required. In the case of non-APHP researchers, the signature of a collaboration contract is moreover mandatory.

**Acknowledgments**

The authors would like to thank the AP-HP data warehouse, which provided the data and the computing power to carry out this study under good conditions. We wish to thank all the medical colleges, including internal medicine, rheumatology, dermatology, nephrology, pneumology, hepato-gastroenterology, hematology, endocrinology, gynecology, infectiology, cardiology, oncology, emergency and intensive care units, that gave their agreements for the use of the clinical data.

**Competing interest**

Authors declare no competing interest

**Consent for publication**

Not applicable

**Funding**

Not applicable

**Authors contribution**

Christel Gérardin: worked on the conceptualization, data curation, formal analysis, investigation, methodology, software, validation, original drafting, writing, revising, and editing the manuscript.

Yuhan Xiong: worked on investigation, methodology, software, validation.

Perceval Wajsbürt: worked on the investigation, the software, the revision of the manuscript.

Fabrice Carrat: worked on conceptualization, methodology, project administration, supervision, writing - original version, writing - revision and editing of the manuscript.

Xavier Tannier: worked on conceptualization, formal analysis, methodology, writing - original version, writing - revision and editing of the manuscript.